\documentclass[final]{nesy2026} 

\usepackage{longtable}

\usepackage{comment} 

\usepackage{booktabs}
\usepackage[load-configurations=version-1]{siunitx} 

\usepackage[shortlabels]{enumitem}

\theorembodyfont{\upshape}
\theoremheaderfont{\scshape}
\theorempostheader{:}
\theoremsep{\newline}

\usepackage{algorithm}
\usepackage[noend]{algcompatible}
\usepackage{listings}
\usepackage{dsfont}  

\title[Adaptive GR(1)-Specification Repair for Liveness-Preserving Shielding]{Adaptive GR(1)-Specification Repair for Liveness-Preserving Shielding in Reinforcement Learning}

 \clearauthor{\Name{Tiberiu-Andrei Georgescu\nametag{\thanks{These authors contributed equally to this work.}$^{1}$}} \Email{tibi.geo@imperial.ac.uk}\\
 \Name{Alexander W. Goodall\nametag{$^{*1}$}} \Email{a.goodall22@imperial.ac.uk}\\
 \Name{Francesco Belardinelli\nametag{$^{1}$}} \Email{francesco.belardinelli@imperial.ac.uk}\\
 \Name{Dalal Alrajeh\nametag{$^{1}$}} \Email{dalal.alrajeh04@imperial.ac.uk}\\
 \Name{Sebastian Uchitel\nametag{$^{2,1}$}} \Email{suchitel@dc.uba.ar}\\
 \addr $^{1}$ Department of Computing, Imperial College London\\
 \addr $^{2}$ University of Buenos Aires, CONICET and Universidad de San Andrés, Argentina
 }

\begin{document}

\maketitle

\begin{abstract}
Shielding is widely used to enforce safety in reinforcement learning (RL), ensuring that an agent's actions remain compliant with formal specifications. Classical shielding approaches, however, are often static, in the sense that they assume fixed logical specifications and hand-crafted abstractions. While these static shields provide safety under nominal assumptions, they fail to adapt when environment assumptions are violated. In this paper, we develop an adaptive shielding framework based on Generalized Reactivity of rank 1 (GR(1)) specifications, a tractable and expressive fragment of Linear Temporal Logic (LTL) that captures both safety and liveness properties. Our method detects environment assumption violations at runtime and employs Inductive Logic Programming (ILP) to automatically repair GR(1) specifications online, in a systematic and interpretable way. This ensures that the shield evolves gracefully, keeping liveness achievable and minimally weakening goals only when necessary. We consider two case studies: Minepump and Atari Seaquest; showing that (i) static symbolic controllers are often severely suboptimal when optimizing for auxiliary rewards, and (ii) RL agents equipped with our adaptive shield maintain near-optimal reward and perfect logical compliance compared with static shields.\renewcommand{\thefootnote}{\textdagger}\footnote{Access to code: \url{https://github.com/sacktock/Adaptive_GR-1-_Shielding}}\renewcommand{\thefootnote}{\arabic{footnote}}
\end{abstract}

\section{Introduction}

Advancements in reinforcement learning (RL) \citep{sutton2018reinforcement} and deep learning \citep{lecun2015deep} have enabled increasingly capable autonomous systems, from robotics to autonomous vehicles and multi-agent search-and-rescue \citep{koos2013fast,chatzilygeroudis2018reset,bongard2006resilient,chen2024end,petrlik2022uavs}. In safety-critical settings, however, optimized behaviour alone is insufficient: agents must also satisfy formal safety constraints, and their behaviour should remain interpretable enough to support diagnosis and trust \citep{camilli2022xs}. 
Shielding is a prominent approach for enforcing such constraints in RL \citep{bloem2015shield,alshiekh2018safe}. A shield is a runtime controller synthesized from a formal specification that monitors the agent's proposed actions and overrides only those that would violate the specification. Unlike safe RL methods based on constrained optimization or penalties, which empirically reduce unsafe behaviour \citep{achiam2017constrained,ray2019benchmarking,thomas2021safe}, shielding can provide correctness guarantees by restricting execution to a winning region of safe states and actions. This minimal-interference property allows the RL agent to continue optimizing its task objective while its choices remain logically admissible.
A key limitation of existing shielding methods is that the environment abstraction used to synthesize the shield is fixed and assumed correct. In many approaches, safety-relevant dynamics are hand-crafted at design time and treated as complete \citep{alshiekh2018safe,bloem2020s,jansen2020safe,konighofer2023online}. When the deployed environment violates these assumptions, due to misspecification or unexpected changes, the shield may become overly conservative, block useful behaviour, or fail to preserve liveness \citep{konighofer2017shield,konighofer2020shield}. Moreover, when assumptions are implicit rather than encoded declaratively, it becomes difficult to diagnose why a specification is unrealizable or why a shield prevents task completion.

\paragraph{Contributions} We propose an adaptive shielding framework based on Generalized Reactivity of rank 1 (GR(1)) specifications \citep{bloem2012synthesis}. GR(1) provides an interpretable assume-guarantee structure for expressing safety and liveness, while retaining polynomial-time synthesis compared with the double-exponential complexity of full LTL synthesis \citep{pnueli1989synthesis}. Our framework makes environment assumptions explicit, synthesizes an initial shield from these assumptions and the desired system guarantees, detects assumption violations at runtime, and repairs the GR(1) specification online using Inductive Logic Programming (ILP). The repaired specification is then re-synthesized into an updated shield.
Our contribution is embedding an ILP-based specification repair procedure \citep{buckworth2023adapting} within a closed-loop online shield-RL system, with formal guarantees that safety and liveness persist across repair events.
Unlike approaches that estimate probabilities over a fixed transition graph \citep{pranger2021adaptive}, our method can alter the logical structure of the environment abstraction itself, producing human-readable explanations of which assumptions or guarantees were weakened and why. The goal is not to learn a full probabilistic model, but to preserve realizability and liveness while weakening the specification only when necessary. We evaluate the framework on Minepump and Atari Seaquest, showing that static symbolic controllers can be highly suboptimal for auxiliary reward optimization, while RL agents equipped with adaptive GR(1) shields maintain logical compliance and near-optimal reward under changing environment dynamics.

\paragraph{Related work} Shielding originates in the \emph{simplex architecture}, where an unreliable controller is supervised by a dependable safety controller \citep{weiss2018towards}. Formal shield synthesis later framed this as a reactive synthesis problem over logical specifications \citep{bloem2015shield,konighofer2017shield,konighofer2020shield}, and was later integrated with deep RL to provide provable safety during learning and execution \citep{alshiekh2018safe}, typically by solving a two-player safety game over a known abstraction, with the shield ensuring that the system remains inside the winning region.
Recent work has begun to address incomplete environment knowledge. Approximate and model-learning shields refine safety-relevant dynamics from data \citep{goodall2023approximate,he2022androids,xiao2023model}, while automata-learning approaches update models during deployment \citep{waga2022dynamic}. These methods, however, typically recover strong guarantees only after sufficient model convergence and do not explicitly maintain realizability throughout adaptation. Adaptive shielding under uncertainty updates transition probabilities in a fixed abstract Markov decision process \citep{pranger2021adaptive}, while parametric approaches adapt bounded parameters within a fixed proof structure \citep{feng2025adaptive}. In contrast, we focus on structural repair of declarative GR(1) specifications: the logical assumptions and, when necessary, guarantees themselves are modified in response to observed violations.

\section{Background}
\label{sec:background}

\textbf{Reinforcement Learning.}
We model agent-environment interaction via the usual \emph{Markov Decision Process} (MDP).  Formally, an MDP is 
a tuple $M = (S, s_0, A, P, R, \gamma)$, where $S$ is a \emph{finite set of states}, with 
\emph{initial state} 
$s_0$, $A$ is a \emph{finite set of actions}, $P: S \times A \to \Delta(S)$ is the \emph{probabilistic transition function}, with $\Delta(S)$ denoting the set of distributions over $S$, $R: S \times A \to \mathbb{R}$ is the \emph{reward function}, and $\gamma$ is the \emph{discount factor}. The agent's \emph{policy} $\pi: S \to \Delta(A)$ is a probabilistic function mapping states to action distributions, with the goal of learning the \emph{optimal policy} $\pi^*$, that maximizes the \emph{expected discounted cumulative reward}: $V^\pi(s) = \mathbb{E}_\pi\left[\sum_{t=0}^{\infty} \gamma^t R(S_t, A_t) \mid S_0 = s \right]$, also called the \emph{value function}. The \emph{Q-function}, similarly defines the policy's expected discounted cumulative reward, from state $s$ taking action $a$, $Q^\pi(s, a) = \mathbb{E}_\pi\left[\sum_{t=0}^{\infty} \gamma^t R(S_t, A_t) \mid S_0 = s, A_0=a \right]$. Q-learning algorithms such as \emph{deep Q-network} (DQN) \citep{mnih2015human}, iteratively approximate and converge to the optimal Q-function, $Q^* = \max_\pi Q^\pi$, thus extracting the optimal policy via, $\pi^* := \arg\max_a Q^*(s, a)$. In contrast, policy-gradient methods, e.g., \emph{Proximal Policy Optimization} (PPO) \citep{schulman2017proximal}, approximate the value function $V^\pi$ of the policy and iteratively update $\pi$ in the direction of the policy-gradient.

\textbf{Linear Temporal Logic.}
Linear Temporal Logic (LTL) \citep{pnueli1977temporal} is a formal language for specifying dynamics properties over reactive systems.
Over a finite set of \emph{atomic propositions} $AP$, LTL formulas are generated by the following syntax,
\begin{equation*}
    \varphi \;::=\; \top \mid \bot \mid p \mid \neg \varphi \mid \varphi\land\varphi \mid X\,\varphi \mid \varphi \;U\; \varphi
\end{equation*}
where $p\in AP$ is an \emph{atomic proposition}, either True ($\top$) or False ($\bot$) from a given state, negation ($\neg$) and conjunction ($\land$) are the familiar propositional operators, next ($X$) and (strong) until ($U$) are temporal operators. 
As usual, disjunction, $ \varphi\lor\psi \equiv \neg(\neg\varphi\land\neg\psi)$, implies, $ \varphi\to\psi \equiv \neg\varphi\lor \psi$, eventually (finally), $F\,\varphi \equiv \top U \varphi$, and always (globally), $G\,\varphi \equiv \neg F\,\neg\varphi$ can be derived in the usual way. Due to space constraints we defer the full LTL semantics to the appendix. However, for notational convenience, we also consider the following past-time operators: yesterday ($Y$) expresses that a formula held in the previous state, and historically ($H$) expresses that a formula has always held in the past.

\textbf{GR(1) Synthesis and Realizability.}
\label{sec:gr1}
For practical synthesis we focus on a specific fragment of LTL: \emph{Generalized Reactivity of rank 1} (GR(1)) \citep{bloem2012synthesis}. GR(1) is an assume-guarantee specification that describes the interaction between two players: the \emph{environment} $\mathbf{e}$ and the \emph{system} $\mathbf{s}$. Written as a pair $\Phi=\langle \mathcal{A}, \mathcal{G} \rangle$, GR(1) specifications provide a logical structure for the environment’s behaviour, the assumptions $\mathcal{A}$, and the desired system properties, the guarantees $\mathcal{G}$. We have (Boolean) variables $ \mathcal V = \mathcal X \cup \mathcal Y$ partitioned into \emph{inputs} $\mathcal X$ (environment $\mathbf e$) and \emph{outputs} $\mathcal Y$ (system/controller $\mathbf s$). A GR(1) specification has the assume-guarantee structure,
\begin{equation}
        \Phi \;=\; \Big(\theta^{\mathbf e} \;\land\; G\,\rho^{\mathbf e} \;\land\; \bigwedge_{i\in I^{\mathbf e}} GF\,J^{\mathbf e}_i\Big)\;\to\;
\Big(\theta^{\mathbf s} \;\land\; G\,\rho^{\mathbf s} \;\land\; \bigwedge_{i\in I^{\mathbf s}} GF\,J^{\mathbf s}_i\Big) \label{eq:realizability}
\end{equation}
where, $\theta^{\mathbf e},\theta^{\mathbf s}$ are  \emph{initial} propositional conditions over $\mathcal V$, $\rho^{\mathbf e},\rho^{\mathbf s}$ are \emph{transition} relations over current/next variables (one-step invariants, wrapped by $G$), $J^{\mathbf e}_i, J^{\mathbf s}_i$ are \emph{justice} (fairness) predicates over current variables (liveness, wrapped by $GF$).

Equivalently, we can write $\mathcal A = \theta^{\mathbf e} \land G\rho^{\mathbf e} \land \bigwedge_{i\in I^{\mathbf e}} GF J^{\mathbf e}_i$ and $\mathcal G = \theta^{\mathbf s} \land G\rho^{\mathbf s} \land \bigwedge_{i\in I^{\mathbf s}} GF J^{\mathbf s}_i$. 
The specification takes the form of a turn-based two-player game $\langle \mathcal{V},\mathcal{X},\mathcal{Y},\theta^{\mathbf{e}},\theta^{\mathbf{s}},\rho^{\mathbf{e}},\rho^{\mathbf{s}},\varphi\rangle$, where $\varphi$ is the fairness formula, $\bigwedge_{i\in I^{\mathbf s}} GF\,J^{\mathbf e}_i\rightarrow\bigwedge_{i\in I^{\mathbf s}} GF\,J^{\mathbf s}_i$. GR(1) synthesis solves the GR(1) game on states $\nu\in 2^{\mathcal V}$ where, (i) the environment $\mathbf e$ picks inputs $x' \in \mathcal{X}$ consistent with $\mathcal A$, and (ii) the system/controller $\mathbf s$ picks outputs $y' \in \mathcal{Y}$ to satisfy $\mathcal G$.
 
We say a GR(1) specification is \emph{realizable} if and only if the \textit{strict realisability formula}, i.e., Eq.~\eqref{eq:realizability}, holds. Furthermore, if a specification is realizable, then there exists a \textit{winning strategy} $\sigma$ for the system $\mathbf s$ to ensure the fairness formula. This strategy is encoded through a symbolic controller in the form of a fairness-free Fair Discrete System (FDS), which can be synthesized in polynomial-time \citep{bloem2012synthesis}. The set of states from which $\mathbf s$ has a \emph{strategy} $\sigma$ ensuring the fairness formula $\varphi$ is called the \emph{winning region} $\mathcal W \subseteq 2^{\mathcal V}$. 
It is important to emphasize the role of environment assumptions in this process. The assumptions $\mathcal{A}$ represent prior knowledge or design-time modelling of the environment. The controller $\mathbf s$ (or shield) relies on these assumptions to guarantee the system properties $\mathcal{G}$ \citep{bloem2012synthesis}. In practical terms, this means that if an assumption is violated at runtime, the shield’s behaviour is undefined with respect to the original specification.

\begin{example}[Minepump] \label{ex:minepump} The minepump system is a classic reactive-synthesis benchmark \citep{kramer1983conic,joseph1996real} for a controller that prevents flooding while avoiding methane ignition hazards. The controller manages a pump that extracts water from the mine. The guarantees $\mathcal{G}$ require the pump to run when water is high but remain off when methane is present, since pumping under methane risks an explosion. The assumptions $\mathcal{A}$ capture the environment: water gradually decreases while pumping, and, for simplicity, methane and high water are assumed not to co-occur. Formally, the variables are: $\mathcal X=\{\texttt{highwater},\texttt{methane}\}$ (environment) and $\mathcal Y=\{\texttt{pump}\}$ (system). The initial spec is given as follows,
\begin{align*}
\textbf{guarantee1:} \quad & G\big(\texttt{highwater}\to X(\texttt{pump})\big) \\
\textbf{guarantee2:} \quad & G\big(\texttt{methane}\to X(\neg \texttt{pump})\big) \\
\textbf{assumption1:} \quad & G\big((Y(\texttt{pump}) \land \texttt{pump}) \to X(\neg \texttt{highwater})\big) \\
\textbf{assumption2:} \quad & G\big(\neg \texttt{highwater}\lor \neg \texttt{methane}\big)
\end{align*}
The winning region $\mathcal{W}$ consists of all states (valuations of $\mathcal{X}$) from which a winning strategy exists. Since, in the winning region $\mathcal{W}$, $\texttt{highwater}$ and $\texttt{methane}$ cannot be simultaneously true, the system $\mathbf{s}$ always has a strategy,
\begin{align*}
    &\text{If $\texttt{methane}$ then set $\texttt{pump}=0$}\\
    &\text{If $\texttt{highwater}$ then set $\texttt{pump}=1$.}\\
    &\text{Else set $\texttt{pump}=0$.}
\end{align*}
which keeps the system within the winning region $\mathcal{W}$. Notice how this strategy is fixed (e.g., static symbolic controller). Lifting this strategy to the full winning region $\mathcal{W}$ gives us a maximally permissive shield, allowing the pump to be applied when $\texttt{highwater}$ is not true, which is useful when optimizing an auxiliary reward (e.g., variable pricing). However, if $\texttt{highwater}$ and $\texttt{methane}$ were to both evaluate to true simultaneously during execution (violating \textbf{assumption2}), the shield's behaviour would be undefined, as we have left the winning region $\mathcal{W}$, highlighting how brittle this initial specification is. 
\end{example}

\section{The \textsc{RepairRL} Shielding Approach}

\begin{figure}[h]
    \centering
    \includegraphics[width=0.4\linewidth]{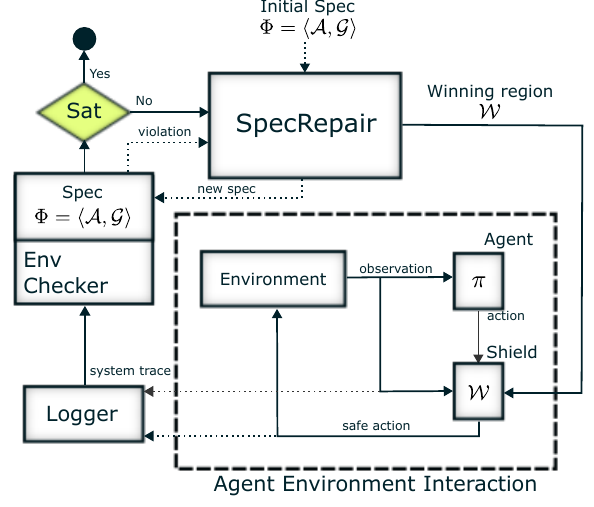}
    \caption{RepairRL: A shield repair framework for logical constraints in GR(1).}
    \label{fig:framework}
\end{figure}

Our overall approach is a hybrid \emph{reactive synthesis} + \emph{reinforcement learning} framework called \textsc{RepairRL} (see Figure~\ref{fig:framework}). The main components are: (1) an RL agent training to maximise rewards provided by the environment; (2) a reactive shield synthesized from a GR(1) specification of the environment assumptions and system guarantees that enforces constraints on the agent's actions; (3) an \emph{Environment Checker} that monitors the traces of the combined system (RL agent + environment) for any violation of the environment assumptions; (4) a \emph{SpecRepair} module that uses ILP, as described in \citep{buckworth2023adapting}, to adapt the specification (learn new assumptions and guarantees, then re-synthesize the shield) when a violation is detected. We describe each aspect of our approach below, including concrete details explaining the division of responsibilities between the learning agent and the synthesized shield.

\subsection{GR(1) Winning Region as Shield}

We remember a GR(1) specification may be synthesized as a symbolic controller. This can take the form of a fairness-free FDS, which we define as
    $\mathcal{D} = \langle\mathcal{V}, \theta, \rho\rangle$,
where $\mathcal{V}=\mathcal{X}\bigcup\mathcal{Y}$ is defined as before, $\theta:2^\mathcal{V}\rightarrow\{True,False\}$ is an \textit{initial condition}, and $\rho:2^\mathcal{V}\times2^\mathcal{V}\rightarrow\{True,False\}$ is a \textit{transition relation}. 
As stated in \citep{bloem2012synthesis}, the fairness-free FDS is a symbolic representation of the Mealy machine, or finite-state reactive system, for those familiar with shielding literature \citep{alshiekh2018safe}.

Given a specification $\Phi$, if Eq.~\eqref{eq:realizability} holds, we may then extract a \textit{maximally permissive safety-shield} as the fairness-free FDS $\mathsf{Sh} = \langle\mathcal{V}, \hat{\theta}, \hat{\rho}\rangle$, where $\hat{\theta}=\theta^e\rightarrow(\theta^s\land\mathcal{W})$ and $\hat{\rho}=(\mathcal{W}\land\rho^e)\rightarrow\rho^s$. A symbolic controller implementing $\mathsf{Sh}$ has the following property.
\begin{proposition}[Shield Closure under the Winning Region]
\label{prop:shield} For a given realizable specification $\Phi$ synthesized as a fairness-free FDS $\mathsf{Sh} = \langle\mathcal{V}, \hat{\theta}, \hat{\rho}\rangle$, a run of consecutive states $(\nu_0,\nu_1,\nu_2,\dots)\in (2^{\Sigma})^\omega$ that satisfies the initial environment condition $\nu_0\models\theta^e$ and transition relation $\forall i\in \mathbb{N}. (\nu_i,\nu_{i+1})\models\rho^e$, will belong to the winning region $\forall i\in \mathbb{N}.\nu_i\in\mathcal{W}$.
\end{proposition}
The aforementioned ensures that, for every state $\nu$ belonging to a run, there is always a strategy $\sigma$ that can satisfy the fairness goal $\varphi$, as long as our environment behaves as expected. This allows our shield to act both towards \textit{safety enforcement} and \textit{liveness preservation}, stopping actions that may force the system into a \emph{liveness-violating ergodic component} (i.e., an infinite loop that never visits at least one of the goal states).

\subsection{Integration with Reinforcement Learning}


To integrate RL agents, based on the MDP formulation, with logic-based shields, we need a concrete way to map MDP states to the Boolean predicates used by the shield. We therefore introduce the following labelling functions:
\begin{quote}
    $\bullet$ \emph{State labelling function:} $L_{\{S\}} : S \to 2^{\mathcal{X}}$ assigns to each state $s \in S$ of the MDP a valuation over the environment inputs $\mathcal{X}$ of the GR(1) assumptions.\\
$\bullet$ \emph{Action labelling function} $L_{\{A\}} : A \to 2^{\mathcal{Y}}$ maps each concrete agent/MDP action $a \in A$ to a valuation over the Boolean output variables $\mathcal{Y}$ used in the GR(1) guarantees.
\end{quote}

\textbf{Definition of Safe Action.} Correctness, with respect to a GR(1) specification, is split into two parts: \emph{safety} and \emph{liveness}. In the context where all environment assumptions are satisfied, a system action is considered {\em correct} if (1) it adheres to the safety invariants $G \rho_s$ and (2) it does not let the environment force the violation of any liveness property $GF J^{\mathbf s}_i$.

\textbf{Derived Safe Action Set.} At runtime the set of \emph{allowable (correct) actions} from MDP state $s$ and shield state $\nu$ is formally defined as,
\begin{equation*}
    A^{\mathsf{Sh}}(s, \nu) = \{ a \in A \mid \big(\nu, (L_{\{A\}}(a),L_{\{S\}}(s))\big)\models \hat{\rho} \}
\end{equation*}

The shield $\mathsf{Sh}$, thus only permits actions $a \in A^{\mathsf{Sh}}(s, \nu)$. If the agent proposes an action outside this set, the shield overrides it with a permissible alternative, either using an arbitrary partial ordering of the actions $A$ (e.g., via Q-function), or randomly sampling from $A^{\mathsf{Sh}}(s, \nu)$. This guarantees that the joint execution trace respects the guarantees $\mathcal{G}$ of the GR(1) specification, provided the environment adheres to the assumptions $\mathcal{A}$. This is formalized as follows.
\begin{proposition}[Correctness and Minimal Deviation] \label{prop:minimaldeviation} Let $\Phi= \langle \mathcal{A}, \mathcal{G} \rangle$ be realizable with winning region $\mathcal{W}$, and assume the environment satisfies $\mathcal{A}$. (i) The shield $\mathsf{Sh}$ synthesized from $\Phi$ ensures that the safe action set $A^{\mathsf{Sh}}(s,\nu)$ is non-empty for every reachable state $(s, \nu)$ from the initial conditions $(s_0, \nu_0)$. Hence, the agent's execution is deadlock-free and satisfies the safety invariants consistent with $\rho^s$. (ii) The shield deviates from the policy $\pi$ if and only if the next state transitions out of the winning region $(\nu,L_{S}(s),(L_{A}(\pi(s)))) \not\models \hat{\rho}$. Thus, liveness remains \emph{achievable} from every visited state $(s, \nu)$ (there exists a strategy to realize $\bigwedge_{i\in I^{\mathbf s}} GF\,J^{\mathbf s}_i$), although it is strictly not enforced by $\mathsf{Sh}$.
\end{proposition}

\subsection{Online Shield Adaptation}

We now detail the full online shield adaptation procedure, starting from initial shield synthesis, monitoring, and specification repair and re-synthesis. This is deployed in-the-loop with the RL agent during both training, so as to learn an optimal policy under the restricted action space, and during deployment for correctness guarantees and adaptation in the face of environment violations; the shield ensures correctness by filtering out precisely those actions that would violate one-step safety or leave $\mathcal{W}$, any adaptations (assumption/guarantee weakening) are made explicit.

\textbf{Initial Shield Synthesis.} Before deployment, we assume that the system engineer or domain expert provides an initial formal specification $\Phi = \langle \mathcal{A}, \mathcal{G} \rangle$, that captures what the system is supposed to do and what environment assumptions it relies on. We then use a GR(1) synthesis tool (such as Spectra \citep{maoz2021spectra}) to compute the winning region $\mathcal{W}$ of the safety game between environment $\mathbf{e}$ and system $\mathbf{s}$. From this, we can derive the set of allowed (safe) actions, given the agent's trajectory through the environment. This set of allowed actions is used to filter the agent’s choices.  

\textbf{Deployment and Monitoring.} Once the RL agent and the initial shield are deployed, at each timestep the \emph{Environment Checker} observes the environment’s state and transitions. If it detects any event that violates the environment assumptions $\mathcal{A}$, it raises a flag indicating an environment failure. For example, if both $\texttt{methane}$ and $\texttt{highwater}$ are simultaneously true, this violates \textbf{assumption2} in the initial spec of the Minepump example. At the moment of detecting a violation, the system’s formal guarantees may no longer hold, and continuing operation with the old shield could lead to an imminent safety failure. We then pause the agent’s normal operation and initiate the \emph{SpecRepair} module. 

\textbf{Specification Repair.} The goal of the \emph{SpecRepair} module is to produce a new realizable specification $\Phi' = \langle \mathcal{A}', \mathcal{G}' \rangle$ that accounts for the observed environment behaviour and is realizable under the current violating trace $\tau$. We employ a counterexample-guided inductive learning approach, which follows a ``learner-oracle'' framework \citep{jha2010oracle}. In our implementation, inspired by the algorithm proposed by Buckworth et al. in \citep{buckworth2023adapting}, the learner is an ILP learner (we use ILASP \citep{law2014ilasp}) and the oracle is a GR(1) synthesizer (Spectra \citep{maoz2021spectra}). A higher level overview of the repair algorithm is outlined in Algorithm \ref{alg:repair}, for interested readers we provide a more complete breakdown of the algorithm in the appendix. We also refer the reader to \citep{buckworth2023adapting} for further details regarding how the ILP learner weakens guarantees.

\begin{algorithm}[h]
\caption{Specification Repair (\emph{SpecRepair})}
\label{alg:repair}
\begin{algorithmic}[1]
\Statex \textbf{Input:} Violating trace $\tau$, current spec $\Phi = \langle \mathcal{A}, \mathcal{G} \rangle$
\Statex \textbf{Output:} Realizable spec $\Phi' = \langle \mathcal{A}', \mathcal{G} \rangle'$
\STATE $A' \gets \text{learner.weaken\_assumptions}(A, \tau)$
\IF{$\text{oracle.is\_realizable}(\langle A', G \rangle)$}
\STATE \textbf{return} $\text{oracle.get\_controller}(\langle A', G \rangle)$
\ELSE
\STATE $cs \gets \text{oracle.counter\_strategy}(\langle A', G \rangle)$
\STATE $G' \gets \text{learner.weaken\_guarantees}(A', G, \tau, cs)$
\STATE \textbf{return} $\text{oracle.get\_controller}(\langle A', G' \rangle)$
\ENDIF
\end{algorithmic}
\end{algorithm}

\textbf{Soundness.} This adaptation happens online and autonomously, without manual specification rewriting. Since the repair is based on formal counterexamples and logic-based learning, the resulting specification $\Phi' = \langle \mathcal{A}', \mathcal{G}' \rangle$ is sound, which we formalise below.
\begin{theorem}[Soundness of SpecRepair] \label{thm:soundness}
Given a realizable specification $\Phi = \langle \mathcal{A}, \mathcal{G} \rangle$ and a violating trace $\tau \not\models \mathcal{A}$, the SpecRepair procedure produces a new specification $\Phi' = \langle \mathcal{A}', \mathcal{G}' \rangle$ such that:
\begin{quote}
(i) $\tau \models \mathcal{A}'$ (the violating trace is consistent with the repaired assumptions). \\
(ii) $\Phi'$ is realizable, i.e., there exists a strategy for the system ensuring all guarantees $\mathcal{G}'$ under all environment behaviours satisfying $\mathcal{A}'$. \\
(iii) $\forall \sigma \in \Sigma^\omega$, $\sigma\models\mathcal{A}\rightarrow\sigma\models\mathcal{A}'$ (any infinite trace formerly allowed by the previous set of assumptions is allowed by the new set of assumptions. In other words, the new set of assumptions is equivalent or weaker to the old one). \\
(iv) $\forall \sigma \in \Sigma^\omega$, $\sigma\models\mathcal{G}\rightarrow\sigma\models\mathcal{G}'$ (equivalent to property (3), but for the set of guarantees).
\end{quote}
\end{theorem}
\begin{corollary} \label{cor:deadlockfree} From (i) and (iv): the winning region $\mathcal{W}'$ induced by $\Phi'$ contains the current execution state reached by $\tau$. This ensures that the agent remains \emph{deadlock-free} after repair.
\end{corollary}
Hence, the repaired shield synthesized from $\Phi'$ both enforces $\mathcal{G}'$ under $\mathcal{A}'$ and guarantees safe continuation of the ongoing execution without interruption. Furthermore, the use of GR(1) also keeps the computational burden manageable for real-time adaptation.

\section{Experimental Evaluation}

We evaluate whether adaptive GR(1) repair preserves logical compliance under design--deployment mismatch while retaining the reward benefits of RL. We consider two domains: Minepump, a compact stochastic MDP based on a standard reactive-synthesis benchmark (c.f., Example \ref{ex:minepump}), and Atari Seaquest, a pixel-based RL task with safety-critical oxygen constraints. In both domains, agents are trained under the nominal specification and evaluated under perturbed dynamics that violate an environment assumption. Further environment details, specifications, hyperparameters, and repair traces are given in Appendix~\ref{sec:appendix-eval}.

We report training reward, training success, evaluation reward, evaluation success, and the evaluation override rate. Success denotes satisfaction of the repaired target guarantees, while override rate measures the fraction of timesteps at which the shield replaces the agent's proposed action. For the standard RL baselines use Stable Baselines3 \citep{stablebaselines3} with shared hyperparameters within each domain (c.f., Appendix \ref{sec:hyperparameters}). Using MASA-Safe-RL \citep{Goodall2025MASASafeRL} we also include two standard safe RL baselines, constrained policy optimization (CPO) \citep{achiam2017constrained} and PPO Lagrangian \citep{ray2019benchmarking}, as part of the evaluation both enforce safety through constrained policy optimization rather than action intervention.

\paragraph{Minepump.}
Minepump tests whether a shield can recover from an incorrect logical assumption. The nominal specification assumes that $\texttt{highwater}$ and $\texttt{methane}$ never occur simultaneously. During deployment this assumption may be violated, creating states in which the original guarantees conflict: the pump should be switched on after high water, but switched off after methane. Our adaptive shield detects this violation, repairs the specification by conditioning the pumping requirement on $\neg\texttt{methane}$, and re-synthesizes the shield online. Additionally, we encode a liveness objective requiring high water to always be eventually cleared. This is represented in GR(1) by raising a system flag whenever $\texttt{highwater}$ occurs and adding the justice guarantee $GF(\neg\texttt{highwater\_flag})$. The resulting specification is unrealizable under the original assumptions: the environment could keep $\texttt{methane}$ true indefinitely, preventing the pump from ever running. We therefore add the environment justice assumption $GF(\neg\texttt{methane}\land X(\neg\texttt{methane}))$, which guarantees a two-step methane-free window in which water can be cleared. This illustrates an important feature of our approach: realizability checks expose the liveness assumptions needed for the shield to be correct, rather than leaving them implicit in the controller.

We compare against two static symbolic controllers \citep{buckworth2023adapting}: Static Ctrl.~1 is synthesized from the nominal spec, whereas Static Ctrl.~2 is synthesized from the repaired spec. We also compare to unshielded DQN \citep{mnih2015human}, and DQN agents equipped with static shields \citep{alshiekh2018safe} that either enforce weakened subsets of the guarantees (\textbf{Static1/2}) or pass actions through in deadlock states (\textbf{Static*}). Table~\ref{tab:minepump-results} shows that the nominal symbolic controller is compliant during training but fails completely in deployment. The repaired symbolic controller remains compliant, but obtains substantially lower reward, illustrating the cost of a fixed symbolic policy. Unshielded DQN and the single-guarantee static shields achieve higher reward but fail the logical task. The static shield that enforces both guarantees succeeds during training, but fails in deployment once assumption-violating deadlock states occur. By contrast, our adaptive shield achieves perfect compliance in both training and deployment while retaining rewards comparable to the strongest static-shielded RL baseline. The additional safe RL baselines further clarify the role of online symbolic adaptation. Although CPO and PPO-Lag satisfy the constraints during training, both fail to generalize once the nominal environment assumption is violated.

\begin{table*}[!t]
\tiny
\centering
\caption{Minepump results after 200K timesteps (10 seeds with mean $\pm$ SE).}
\label{tab:minepump-results}
\begin{tabular}{lccccc}
\toprule
\textbf{Controller / Agent} & 
\textbf{Train Rew.} & 
\textbf{Train Succ.} & 
\textbf{Eval Rew.} & 
\textbf{Eval Succ.} & 
\textbf{Eval Ovrr.} \\
\midrule
Static Ctrl.~1 & -849.64 (std: 36.92)   & 1.00 (std: 0.00) & -821.10  (std: 56.26) & 0.00 (std: 0.00) & -- \\
Static Ctrl.~2  & -827.87 (std: 51.79)   & 1.00 (std: 0.00) & -1018.29 (std: 66.36)  & 1.00 (std: 0.00) & -- \\
DQN  &  -883.10 $\pm$ 1.72 & 0.00 $\pm$ 0.00 & -672.36 $\pm$ 6.47 & 0.00 $\pm$ 0.00 & -- \\
DQN + \textbf{Static1}  &   -882.25 $\pm$ 1.41 & 0.00 $\pm$ 0.00 & -669.16 $\pm$ 6.90 & 0.00 $\pm$ 0.00 & 0.01 $\pm$ 0.00 \\
DQN + \textbf{Static2} & -816.33 $\pm$ 2.70 & 0.00 $\pm$ 0.00 & -813.90 $\pm$ 16.43 & 0.00 $\pm$ 0.00 & 0.40 $\pm$ 0.00 \\
DQN + \textbf{Static*} & -806.73 $\pm$ 1.83 & 1.00 $\pm$ 0.00 & -747.00 $\pm$ 10.02 & 0.00 $\pm$ 0.00 & 0.40 $\pm$ 0.00 \\
DQN + \textbf{Adaptive (ours)} & -807.21 $\pm$ 1.66 & 1.00 $\pm$ 0.00 & -764.98 $\pm$ 6.68 & 1.00 $\pm$ 0.00 & 0.19 $\pm$ 0.05 \\
CPO & -845.30 $\pm$ 22.10 & 1.00 $\pm$ 0.00  & -780.30$\pm$ 45.03 & 0.01 $\pm$ 0.00 & -- \\
PPOLag  & -754.77 $\pm$ 54.01 & 1.00 $\pm$ 0.00 & -691.00 $\pm$ 36.01 & 0.01 $\pm$ 0.00 & -- \\
\bottomrule
\end{tabular}
\end{table*}

\paragraph{Seaquest.}
Seaquest is used to evaluate our approach in a high-dimensional visual domain. The initial GR(1) abstraction assumes that oxygen depletes by one unit per timestep while the submarine is underwater. In the evaluation environment, this assumption is violated: once oxygen reaches a fixed threshold, oxygen begins depleting twice as fast. The adaptive shield repairs the oxygen-transition assumption and weakens only those reachability obligations that are no longer realizable, while preserving the safety guarantee that the submarine never runs out of oxygen. Seaquest also contains liveness-style reachability obligations: when a diver appears at a depth segment, the system should eventually reach that segment while maintaining the safety guarantee that oxygen never reaches zero. Under the nominal oxygen model, these obligations are realizable. However, once oxygen depletes faster than assumed, some depth-reachability goals become unrealizable without violating oxygen safety. The adaptive repair therefore preserves the safety guarantee and weakens only the affected reachability obligations.
As the RL backbone we use PPO \citep{schulman2017proximal}, comparing against a na{\"i}ve hand-crafted shield, two static shields \citep{alshiekh2018safe}: \textbf{Static} and \textbf{Repaired} synthesized from the nominal spec and repaired spec respectively, and our adaptive shielding framework. As shown in Table~\ref{tab:seaquest-results}, all shielded agents satisfy the oxygen safety guarantee during training, while unshielded PPO fails in most episodes. Under the perturbed evaluation dynamics, both naive and static shields fail because their winning regions are computed from the incorrect oxygen model. The pre-repaired and adaptive shields both maintain perfect evaluation success with low override rates. Rewards decrease in the perturbed environment, reflecting both the harder dynamics and limited transfer from the nominal emulator, but the adaptive shield preserves compliance without requiring the repaired model to be known. CPO and PPO-Lag achieve perfect oxygen-safety success in both training and evaluation, but only by learning highly conservative policies with substantially lower reward than the shielded PPO agents.

\begin{table*}[!t]
\centering
\tiny
\caption{Seaquest results after 10M timesteps (5 seeds with mean $\pm$ SE).}
\label{tab:seaquest-results}
\begin{tabular}{lccccc}
\toprule
\textbf{Controller / Agent} & 
\textbf{Train Rew.} & 
\textbf{Train Succ.} & 
\textbf{Eval Rew.} & 
\textbf{Eval Succ.} & 
\textbf{Eval Ovrr.} \\
\midrule
PPO    &  142.48 $\pm$ 3.20 & 0.19 $\pm$ 0.03 & 0.00 $\pm$ 0.00 & 0.00 $\pm$ 0.00 & -- \\
PPO + \textbf{Naive}       &  377.21 $\pm$ 22.95 & 1.00 $\pm$ 0.00 & 0.20 $\pm$ 0.20 & 0.01 $\pm$ 0.01 & 0.18 $\pm$ 0.02 \\
PPO + \textbf{Static}      &  224.03 $\pm$ 21.26 & 1.00 $\pm$ 0.00 & 0.00 $\pm$ 0.00 & 0.00 $\pm$ 0.00 & 0.08 $\pm$ 0.01 \\
PPO + \textbf{Repaired}    &  261.77 $\pm$ 32.91 & 1.00 $\pm$ 0.00 & 75.60 $\pm$ 13.94 & 1.00 $\pm$ 0.00 & 0.05 $\pm$ 0.01 \\
PPO + \textbf{Adaptive (ours)}  & 224.58 $\pm$ 24.87 & 1.00 $\pm$ 0.00 & 58.80 $\pm$ 20.42 & 1.00 $\pm$ 0.00 & 0.06 $\pm$ 0.01 \\
CPO & 20.00 $\pm$ 0.00 & 1.00 $\pm$ 0.00 & 20.00 $\pm$ 0.00 & 1.00 $\pm$ 0.00 & -- \\
PPOLag  & 20.13 $\pm$ 10.05 & 1.00 $\pm$ 0.00 & 22.45 $\pm$ 12.40 & 1.00 $\pm$ 0.00 & -- \\
\bottomrule
\end{tabular}
\end{table*}

\section{Conclusion}
Our results show that adaptive GR(1) specification repair is a practical approach for maintaining deep RL agents both safe and effective when environment assumptions are violated at runtime. By deploying agents with an \emph{Environment Checker} and invoking the ILP-guided \emph{SpecRepair} module only upon detected assumption violations, we can re-synthesize shields online with transparent, human-readable changes to assumptions and guarantees. The explicitness of our approach improves both interpretability and trust, as the assumptions that underpin realizability are no longer hidden in the synthesized controller but are exposed, and when necessary, weakened minimally and visibly.
Two limitations point to natural extensions. First, our specification formalism relies on LTL in GR(1) form, which requires a finite, discrete state abstraction; while this worked well for propositions such as oxygen-level or depth in Atari Seaquest, domains with rich continuous dynamics may resist clean logical discretization, motivating future work on more expressive logics such as Signal Temporal Logic or hybrid discrete-continuous formalisms. Second, although GR(1) synthesis itself is efficient, the iterative learner-oracle loop (e.g., ILASP + Spectra) can become a computational bottleneck for large specifications or frequent repairs; faster repair algorithms would help scale adaptive shielding to more complex tasks and longer deployments. Extending our approach to adversarial or multi-agent systems via more expressive fragments of LTL, such as GR(k), is a longer-term goal.
%
%

\acks{The research described in this paper was partially supported by the EPSRC (grant number EP/X015823/1) and by the UK Research and Innovation (grant number EP/S023356/1) in the UKRI Centre for Doctoral Training in Safe and Trusted Artificial Intelligence (\url{www.safeandtrustedai.org}).}

\bibliography{nesy2026-sample}

\newpage
\appendix

\section{LTL Semantics}

LTL allows us to write properties about sequences of states (or traces) and constrains the infinite execution trace of a system, Let $\tau = s_0 s_1 s_2 \ldots$ be an infinite word over $2^{AP}$ and let $(\tau, i) \models \varphi$ denote satisfaction at position $i\ge 0$. Then the semantics of LTL formulas is given as follows:
\begin{align*}
    (\tau, i) &\models p & \text{iff} & \ \ \ \   p \in s_i \; \text{(for $p\in AP$)}\\
(\tau, i) &\models \neg\varphi & \text{iff} & \ \ \ \  \neg(\tau, i) \models \varphi\\
 (\tau, i) &\models \varphi\land\psi & \text{iff} & \ \ \ \  (\tau, i) \models \varphi \land (\tau, i) \models \psi\\
(\tau, i) &\models X\,\varphi & \text{iff} & \ \ \ \ (\tau, i{+}1) \models \varphi \\
(\tau, i) & \models \varphi\,U\,\psi & \text{iff} & \ \ \ \  \exists k\ge i.\; (\tau,k) \models \psi \\
&&&\;\land\; \forall j\in[i,k)\!:\; (\tau,j) \models \varphi
\end{align*}
Formally, the semantics of the past time operators yesterday ($Y$) and historically ($H$) are defined as follows:
\begin{align*}
    (\tau, i) &\models Y\,\varphi & \text{iff} & \ \ \ \ (i > 0) \land (\tau, i{-}1) \models \varphi\\
    (\tau, i) &\models H\,\varphi & \text{iff} & \ \ \ \ \forall j < i.\; (\tau,j) \models \varphi
\end{align*}
While full LTL is an expressive language (e.g., safety, liveness), controller synthesis from arbitrary LTL formula is doubly-exponential in the size $|\varphi|$ of the formula.

\section{Proofs}

\subsection{Proof (sketch) for Proposition \ref{prop:shield}}

Let $\Phi$ be realizable. By Theorem 3.1 of \cite{bloem2012synthesis}, the GR(1) game admits a winning strategy $\sigma$ for the system, and the winning region $\mathcal{W} \subseteq 2^\mathcal{V}$ is the greatest fixpoint of the standard GR(1) predecessor construction. In particular:
\begin{itemize}
    \item $\mathcal{W}$ is \emph{closed under the strategy} $\sigma$.
    \item From every state in $\mathcal{W}$, $\sigma$ enforced satisfaction of the fairness condition.
    \item $\mathcal{W}$ is maximal among such closed sets.
\end{itemize}
The shield $\mathsf{Sh}$ is constructed as the fairness-free FDS:
\begin{equation*}
    \hat{\theta} = \theta^{\mathbf e} \rightarrow (\theta^{\mathbf s} \land \mathcal{W}),
\qquad
\hat{\rho} = (\mathcal{W} \land \rho^{\mathbf e}) \rightarrow \rho^{\mathbf s}.
\end{equation*}
Thus:
\begin{enumerate}[(1)]
    \item \textbf{Initialization.} If $\nu_0 \models \theta^{\mathbf e}$, then by construction of $\hat{\theta}$ we must have $\nu_0 \models \mathcal{W}$, hence $\nu_0 \in \mathcal{W}$.
    \item \textbf{Inductive step.} Suppose $\nu_i \in \mathcal{W}$. Since the environment respects $\rho^{\mathbf e}$ and $\sigma$ is winning from every state in $\mathcal{W}$ \citep{bloem2012synthesis}, there exists a system move satisfying $\rho^{\mathbf s}$ such that the successor state remains in $\mathcal{W}$. Because $\hat{\rho}$ enforces precisely these $\sigma$-consistent transitions, it follows that $\nu_{i+1} \in \mathcal{W}$.
\end{enumerate}
The proof follows from induction on $i$.

\subsection{Proof (sketch) for Proposition \ref{prop:minimaldeviation}}

By \cite{bloem2012synthesis} we have:
\begin{itemize}
    \item the winning region $\mathcal{W}$ is the \emph{maximal} set of states from which the system has a winning strategy.
    \item If a state $\nu \not \in \mathcal{W}$, then no strategy can ensure satisfaction of $\Phi$ from $\nu$.
\end{itemize}
We define the safe action set as follows,
\begin{equation*}
    A^{\mathsf{Sh}}(s, \nu) = \{ a \in A \mid \big(\nu, (L_{\{A\}}(a),L_{\{S\}}(s))\big)\models \hat{\rho} \}
\end{equation*}
By construction of $\hat \rho$ and $A^{\mathsf{Sh}}(s, \nu)$, the shield allows exactly those transitions that preserve $\mathcal{W}$.
Suppose, for a contradiction, that the shield blocks some action $a$ from $\nu \not \in \mathcal{W}$ that preserves $\Phi$. Then taking $a$ would lead to some successor in $\nu' \in \mathcal{W}$. But this contradicts the definition of $\hat{\rho}$, which allows all $\mathcal{W}$-preserving transitions.
Conversely, any action leading to $\nu' \not \in \mathcal{W}$ must be blocked, since from such a state no winning strategy exists.

\subsection{Proof (sketch) for Theorem \ref{thm:soundness}}

Algorithm \ref{alg:repair} performs iterative weakening (Line 6) using the repair procedure of \cite{buckworth2023adapting}. From \cite{buckworth2023adapting} we have:
\begin{itemize}
    \item Every weakening step produces $\Phi'$ such that:
    \begin{enumerate}[(1)]
        \item $\mathcal{A}'$ admits the observed violating trace $\tau$,
        \item If realizable, the GR(1) synthesis procedure returns a winning strategy.
    \end{enumerate}
\end{itemize}
GR(1) synthesis correctness \citep{bloem2012synthesis} ensures that:
\begin{itemize}
    \item If strict realizability holds, a fairness-free FDS implementing $\Phi'$ is returned.
    \item This controller enforces $\Phi'$ for all assumption-compliant runs.
\end{itemize}
Thus, each repair iteration either:
\begin{itemize}
    \item Produces an unrealizable candidate (rejected via counter strategy), or
    \item Produces a realizable $\Phi'$ with a correct shield.
\end{itemize}
Therefore, the final $\Phi'$ returned by Algorithm \ref{alg:repair} is enforced by its synthesized shield. 

\subsection{Proof (sketch) for Corollary \ref{cor:deadlockfree}}

For a contradiction, assume the repaired shield allows a current run $\tau$ that violates $\Phi'$. Since $\Phi'$ was declared realizable synthesis produced a winning strategy $\sigma'$ for the GR(1) game \citep{bloem2012synthesis}. By Theorem \ref{thm:soundness}, the fairness-free FDS extracted from $\sigma'$ implements $\Phi'$. Thus, every run of the shield consistent with assumptions must satisfy $\Phi'$. If $\tau$ violates $\Phi'$, then either:
\begin{itemize}
    \item The shield deviated from the synthesized FDS -- impossible by construction.
    \item $\Phi'$ was incorrectly declared realizable -- contradicting the soundness of the synthesis and repair loop of \cite{buckworth2023adapting}.
\end{itemize}
Hence no such $\tau$ exists.

\section{Algorithm Breakdown}
\setcounter{algorithm}{1}
\begin{algorithm}[h]
\caption{Specification Repair (\emph{SpecRepair})}
\label{alg:repair2}
\begin{algorithmic}[1]
\Statex \textbf{Input:} Violating trace $\tau$, current spec $\Phi = \langle \mathcal{A}, \mathcal{G} \rangle$
\Statex \textbf{Output:} Realizable spec $\Phi' = \langle \mathcal{A}', \mathcal{G} \rangle'$
\STATE $A' \gets \text{learner.weaken\_assumptions}(A, \tau)$
\IF{$\text{oracle.is\_realizable}(\langle A', G \rangle)$}
\STATE \textbf{return} $\text{oracle.get\_controller}(\langle A', G \rangle)$
\ELSE
\STATE $cs \gets \text{oracle.counter\_strategy}(\langle A', G \rangle)$
\STATE $G' \gets \text{learner.weaken\_guarantees}(A', G, \tau, cs)$
\STATE \textbf{return} $\text{oracle.get\_controller}(\langle A', G' \rangle)$
\ENDIF
\end{algorithmic}
\end{algorithm}

\begin{enumerate}[(1)]
    \item \emph{Weaken assumptions}: At line 1, using the violation trace $\tau$ (the sequence of states/actions leading up to the assumption failure) as a positive example, the learner proposes a revised set of environment assumptions $\mathcal{A}'$ that is weaker than the original $\mathcal{A}$ and permits the new observed behaviour. For example, for the Minepump case study, a trivial way to weaken the violating assumption (\textbf{assumption2}) is to remove it. 
    \item \emph{Realizability check}: At line 2, the oracle (synthesizer) checks whether the system guarantees $\mathcal{G}$ are still realizable under the new assumptions $\mathcal{A}'$. If $\langle \mathcal{A}', \mathcal{G} \rangle$ is realizable, it means the system can still fulfil the original guarantees even with the environment’s new behaviour, so we synthesize this new specification's winning region $\mathcal{W}'$ (line 3).
    \item \emph{Weaken guarantees (if needed)}: If $\langle \mathcal{A}', \mathcal{G} \rangle$ is not realizable, it means that the environment’s deviation has made it impossible for all the original guarantees to be satisfied. In other words, $\mathcal{G}$ is too strict under the new set of assumptions $\mathcal{A}'$. Therefore, at line 5, we extract a \emph{counter-strategy} -- an example of how the environment can force a violation of $\mathcal{G}$. The logic-based learner then uses this counter-strategy to learn $\mathcal{G}'$, a set of weaker system guarantees, that no longer considers the proposed counter-strategy as a failure. At line 6 we present this process as a single-shot task, but actually this is an iterative process, where multiple counter-strategies may be generated to guide the learning process. We note that the procedure explores \emph{weakenings} in a strictly increasing partial order of deviation from the original specification; trivial guarantees such as $G(\top)$ are reachable only if all strictly weaker repairs fail realizability. The learning procedure of \citep{buckworth2023adapting} further exploits minimal unsatisfiable cores extracted from counter strategies to guide the ILP search toward minimally sufficient assumption and goal weakenings. For a more detailed description of how counter-strategies aid in guiding the learning process for the weakened guarantees $\mathcal{G}'$, we refer the reader to \citep{buckworth2023adapting}. In the Minepump case study, after removing \textbf{assumption2}, either \textbf{guarantee1} or \textbf{guarantee2} need to be weakened. Assuming turning the pump on after methane is observed is highly dangerous behaviour, \textbf{guarantee1} is weakened to: $G(\texttt{highwater} \land \neg \texttt{methane} \to X \texttt{pump})$.
    \item \emph{Re-synthesize}: Finally, at line 7, the oracle synthesizes the new winning region $\mathcal{W}'$ from the weakened specification $\langle \mathcal{A}', \mathcal{G}' \rangle$. This new shield now enforces the modified guarantees $\mathcal{G}'$ under the updated assumptions $\mathcal{A}'$.
\end{enumerate}

\section{Additional Experimental Details}
\label{sec:appendix-eval}


\subsection{Minepump Details}
\label{sec:minepump}

\textbf{Environment.} We extend the canonical GR(1) Minepump specification into a stochastic MDP with exogenous disturbances and auxiliary rewards. In the training environment, the original assumption that $\texttt{methane}$ and $\texttt{highwater}$ never co-occur is enforced. In the deployment environment, this assumption may be violated by stochastic transitions, creating a design--deployment mismatch.

Water inflow follows a three-state Markov chain over rates $\{0,0.5,2\}$, producing intermittent bursts of high inflow. Methane follows a two-state geometric process with switching probabilities
$p_{\text{on}\to\text{off}}=0.25$ and $p_{\text{off}\to\text{on}}=1/6$, producing persistent methane episodes. Energy prices follow a two-state Markov chain with persistence $0.7$, yielding low and high tariffs of $1.0$ and $4.0$, respectively. Formally, 
\begin{equation*}
    \mathbf{P} \;=\;
\begin{bmatrix}
0.80 & 0.18 & 0.02\\
0.10 & 0.80 & 0.10\\
0.02 & 0.28 & 0.70
\end{bmatrix}
\end{equation*}
with $\Pr(S_{t+1}=s' \mid S_t=s)=\mathbf{P}_{s,s'}$, denoting the probability of transitioning from state $s$ to $s'$.

The high-water threshold is $H=10$ and the water cap is $W_{\max}=18$. The pump removes $R=6$ units per step, so two consecutive pumping steps clear high water under the nominal dynamics. Rewards combine switching costs, tariff-scaled pumping costs, and a high-water penalty. This creates a trade-off in which reward-optimal behaviour may pump early, for example during low-tariff periods or just after methane clears, while still respecting the logical guarantees. Formally, the water level $W_t$ is updated at each timestep as follows,
\begin{equation*}
    W_{t+1}=\operatorname{clip}\!\left(
   W_t + r_{S_t} - R \cdot\mathds{1}\{A_t=\texttt{pump}\},
   0, W_{\max} \right)
\end{equation*}
\paragraph{Liveness extension.}
We also evaluate a liveness extension requiring that high water is always eventually cleared. To express this in GR(1), we introduce a system variable $\texttt{highwater\_flag}$ and add:
\begin{quote}
    $\bullet$ $G(\texttt{highwater} \to \texttt{highwater\_flag})$.\\
    $\bullet$ $GF(\neg\texttt{highwater\_flag})$.
\end{quote}
Together, these encode that whenever water rises above the threshold, it must eventually be brought down again.
This specification is unrealizable under the original assumptions, since the environment could keep $\texttt{methane}=1$ forever, preventing the pump from operating. We therefore add the environment justice assumption
\[
GF(\neg \texttt{methane} \land X(\neg \texttt{methane})),
\]
which guarantees that methane eventually clears for at least two consecutive steps. This gives the system a realizable window in which to pump and clear high water.

Liveness assumptions are not directly monitorable from finite traces, since they are interpreted over infinite executions. In practice, the environment checker can issue warnings when no two-step methane-free window has occurred for many timesteps. These warnings do not immediately trigger repair, but make explicit the assumptions on which realizability depends.

\paragraph{Baselines.}
\begin{itemize}
    \item \textbf{Static Symbolic Controller 1}: the nominal winning strategy synthesized from the initial specification, which assumes $\texttt{methane}$ and $\texttt{highwater}$ are disjoint.
    \item \textbf{Static Symbolic Controller 2}: the repaired symbolic controller synthesized after removing the disjointness assumption and enforcing pumping only when $\texttt{highwater}\land\neg\texttt{methane}$ holds.
    \item \textbf{DQN}: an unshielded DQN agent.
    \item \textbf{DQN + Static 1}: a shield that drops the high-water pumping guarantee, allowing inaction in high-water states to avoid deadlock.
    \item \textbf{DQN + Static 2}: a shield that drops the methane safety guarantee, permitting pumping under methane to avoid deadlock.
    \item \textbf{DQN + Static (*)}: a shield that enforces both guarantees where possible, but passes the agent's action through in deadlock states.
    \item \textbf{DQN + Adaptive}: our method, which detects assumption violations, repairs the GR(1) specification online, and re-synthesizes the shield.
\end{itemize}

\paragraph{Training Details.} All DQN agents use the Stable Baselines3 implementation \citep{stablebaselines3} with identical hyperparameters reported in Table \ref{tab:DQN-hyperparameters}. CPO and PPOLag agents use the MASA-Safe-RL implementation with hyperparameters reported in Table \ref{tab:CPO-hyperparameters} and \ref{tab:PPOLag-hyperparameters} respectively. Agents are trained for $200{,}000$ timesteps and evaluated over $20$ episodes. Each shield is used both during training and in the perturbed evaluation environment. Each training run takes approximately 3 minutes on the hardware used: NVIDIA GeForce RTX 4070 with 8GB of RAM.

\paragraph{Metrics.}
Minepump success is measured against the repaired target guarantees:
\[
G((\texttt{highwater}\land\neg\texttt{methane}) \to X(\texttt{pump}))
\]
and
\[
G(\texttt{methane}\to X(\neg\texttt{pump})).
\]
We report training reward, training success, evaluation reward, evaluation success, and evaluation override rate. Results are averaged over 10 random seeds, using standard error for RL agents and standard deviation where indicated for symbolic controllers.

\subsection{Seaquest Details}

\begin{figure}[!h]
    \centering
    \includegraphics[width=0.4\linewidth]{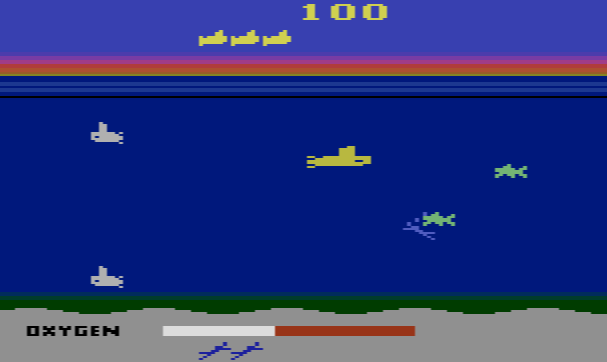}
    \caption{Atari Seaquest: rescue \emph{divers} (blue people), while shooting enemy \emph{sharks} and \emph{submarines} to collect reward.}
    \label{fig:seaquest}
\end{figure}

\paragraph{Environment.} We modify the Atari 2600 Seaquest emulator provided by the Arcade Learning Environment \citep{machado18arcade} by introducing quicker and variable oxygen depletion rates. The task contains a safety-critical resource, the submarine's oxygen supply, and reward-producing objectives such as rescuing divers and avoiding or shooting enemies (c.f., Figure \ref{fig:seaquest}). We introduce faster and variable oxygen depletion in order to test whether the shield can adapt when the engineer's original dynamics model is violated.

\paragraph{Initial abstraction.}
The full GR(1) specification uses Boolean encodings of oxygen and depth. At a high level, the environment variables $\mathcal{X}$ encode the submarine's current $\texttt{oxygen}$ level and $\texttt{depth}$, while the system variables $\mathcal{Y}$ encode movement actions such as $\texttt{up}$ and $\texttt{down}$.

The main environment assumptions are:
\begin{itemize}
    \item \textbf{assumption1}: fixed oxygen depletion,
    \[
    G((\texttt{depth}>0 \land \texttt{oxygen}>0)
    \to X(\texttt{oxygen})=\texttt{oxygen}-1).
    \]
    \item \textbf{assumption2}: deterministic vertical movement, e.g.,
    \[
    G((\texttt{depth}>0 \land \texttt{up})
    \to X(\texttt{depth})=\texttt{depth}-4),
    \]
    with an analogous rule for downward movement.
\end{itemize}

The main system guarantees are:
\begin{itemize}
    \item \textbf{guarantee1}: never run out of oxygen,
    \[
    G(\neg(\texttt{oxygen}=0)).
    \]
    \item \textbf{guarantee2}: eventually reach divers at their depth segment,
    \[
    G(\texttt{diver\_at\_depth}[i] \to F(\texttt{depth}=i)).
    \]
\end{itemize}

The initial shield is synthesized from $\langle\mathcal{A},\mathcal{G}\rangle$ using Spectra \citep{maoz2021spectra}. The maximum oxygen level is $64$ and the maximum depth is $92$. Under the nominal depletion rate of one oxygen unit per timestep, the submarine can still reach divers at any depth while preserving the oxygen safety guarantee.

\paragraph{Violation and repair.}
In evaluation, we violate \textbf{assumption1} when $\texttt{oxygen}=48$: from that point onward, oxygen depletes by two units per timestep rather than one. The adaptive shield weakens the oxygen-transition assumption to allow either depletion rate:
\[
G((\texttt{depth}>0 \land \texttt{oxygen}>0)
\to
(X(\texttt{oxygen})=\texttt{oxygen}-1
\lor
X(\texttt{oxygen})=\texttt{oxygen}-2)).
\]
This repair shrinks the winning region because some depth--oxygen configurations that were previously safe are no longer safe under faster depletion. At the violation threshold, some diver-reaching obligations also become unrealizable at certain depths, so the repair weakens the affected reachability guarantees while preserving \textbf{guarantee1}, the oxygen safety property.

\paragraph{Baselines.}
\begin{itemize}
    \item \textbf{PPO}: unshielded PPO.
    \item \textbf{PPO + Naive}: a static shield that considers only the current oxygen level when enforcing oxygen safety, yielding a suboptimal winning region.
    \item \textbf{PPO + Static}: a static shield synthesized from the initial specification, using both oxygen and depth to implement the full nominal winning region $\mathcal{W}$.
    \item \textbf{PPO + Repaired}: a shield synthesized from the repaired specification in advance, implementing the repaired winning region $\mathcal{W}'$.
    \item \textbf{PPO + Adaptive}: our method, which begins with the nominal shield and switches to the repaired shield only after observing the faster oxygen-depletion violation.
\end{itemize}
\paragraph{Training Details.} All PPO agents use Stable Baselines3 \citep{stablebaselines3} with identical hyperparameters reported in Table \ref{tab:PPO-hyperparameters}. CPO and PPOLag agents use the MASA-Safe-RL implementation with hyperparameters reported in Table \ref{tab:CPO-hyperparameters} and \ref{tab:PPOLag-hyperparameters} respectively. Runs are capped at $10$M timesteps and evaluated over $20$ episodes. The nominal training environment uses an oxygen depletion rate of one, while the evaluation environment contains the assumption violation. Each training run takes approximately 15 hours on the hardware used: NVIDIA GeForce RTX 4070 with 8GB of RAM.

\paragraph{Metrics.} Seaquest success is only measured against \textbf{guarantee1}:
\[
    G(\neg(\texttt{oxygen}=0)),
    \]
as \textbf{guarantee2} is a liveness-type constraint that cannot be enforced. We report training reward, training success, evaluation reward, evaluation success, and evaluation override rate. Results are averaged over 5 random seeds with standard error.

\paragraph{Repair trace.}
The repair trace records why adaptation was triggered and which assumptions or guarantees were changed. For Seaquest, the trace identifies the observed violation of the oxygen-depletion assumption, the repaired oxygen-transition rule, and the resulting weakening of unreachable diver-reachability guarantees. This trace provides an interpretable explanation of how the shield changed and why the repaired specification remains realizable.

\begin{lstlisting}[language={}]
============================================================
Adaptive GR(1) Shield - Repair Trace (Atari Seaquest)
============================================================
[VIOLATION] 
        Assumption violated:
            G(oxygen = 49 -> X(faster_rate = False))
            
	Current state:
	    oxygen = 48, depth = 35, faster_rate = True
------------------------------------------------------------
[STEP 1] Assumption Repair
------------------------------------------------------------
[RESULT] Removed:
    G(oxygen = 49 -> X(faster_rate = False))
------------------------------------------------------------
[STEP 2] Realizability Check
------------------------------------------------------------
[RESULT] Unrealizable.
------------------------------------------------------------
[STEP 3] Guarantee Repair
------------------------------------------------------------
[RESULT] Removed:
    G3: GF(rescues3)
    G4: GF(rescues4)
    -> Realizable
------------------------------------------------------------
[FINAL SPECIFICATION]
------------------------------------------------------------
Assumptions Removed:
  G(oxygen = 49 -> X(faster_rate = False))
  
Assumptions Weakened:
  [None]
  
Guarantees Removed:
  GF(rescues3)
  GF(rescues4)
  
Guarantees Weakened:
  [None]
============================================================
[Result] Success
============================================================
\end{lstlisting}

\paragraph{Full Spec.} The full spec for Atari Seaquest can be found in Appendix \ref{sec:seaquestspec}.
\clearpage

\section{Hyperparameters}
\label{sec:hyperparameters}

\begin{table}[h]
\centering
\caption{SB3 DQN 
hyperparameters for Minepump.}
\label{tab:DQN-hyperparameters}
\begin{tabular}{l c c}
\toprule
\textbf{Name} & \textbf{Symbol} & \textbf{Value}\\
\midrule
Learning rate & $\alpha$ & $1\times10^{-3}$ \\
Replay buffer size & -- & $100{,}000$ \\
Buffer warm-up & -- & $10{,}000$ \\
Batch size & $B$ & $256$ \\
Discount factor & $\gamma$ & $0.99$ \\
Train frequency & -- & every $4$ env steps \\
Target update interval & -- & $1{,}000$ steps \\

$\epsilon$-Greedy exploration  & $\epsilon_{\text{start}},\ \epsilon_\text{final}$ & $1.0 \rightarrow 0.05$ \\
$\epsilon$-Decay steps (fraction) & -- & $0.2$ \\
Network architecture & -- & 2-layer, $[64,64]$ units, ReLU \\
Max gradient norm & -- & $0.5$ \\
\bottomrule
\end{tabular}
\end{table}

\begin{table}[h]
\centering
\caption{SB3 PPO 
hyperparameters for Atari Seaquest.}
\label{tab:PPO-hyperparameters}
\begin{tabular}{l c c}
\toprule
\textbf{Name} & \textbf{Symbol} & \textbf{Value}\\
\midrule
Number of envs & -- & 8 \\
Rollout steps (per env) & -- & $128$ \\
Batch size & $B$ & $256$ \\
PPO epochs & -- & $3$ \\
Discount factor & $\gamma$ & $0.99$ \\
GAE lambda & $\lambda$ & $0.95$ \\
Entropy coefficient & $\beta_{\mathrm{ent}}$ & $0.01$ \\
Value-loss coefficient & $\beta_V$ & $1.0$ \\
Learning rate & $\alpha$ & $2.5\times10^{-4}$ (linear schedule) \\
Clip range & $\epsilon_{\text{clip}}$ & $0.1$ (linear schedule) \\
Network architecture & -- & Nature CNN \citep{mnih2015human}\\
Max gradient norm & -- & $0.5$ \\
\bottomrule
\end{tabular}
\end{table}
\clearpage

\begin{table}[H]
\centering
\caption{MASA CPO hyperparameters.}
\label{tab:CPO-hyperparameters}
\begin{tabular}{l c c}
\toprule
\textbf{Name} & \textbf{Symbol} & \textbf{Value}\\
\midrule
Learning rate & $\alpha$ & $1\times10^{-4}$ \\
Rollout steps & -- & $20{,}000$ \\
Critic updates & -- & $20$ \\
Discount factor & $\gamma$ & $0.99$ \\
GAE lambda & $\lambda$ & $0.9$ \\
Normalize reward advantages & -- & \texttt{True} \\
Normalize cost advantages & -- & \texttt{True} \\
Entropy coefficient & $\beta_{\mathrm{ent}}$ & $0.0$ \\
Value-loss coefficient & $\beta_V$ & $1.0$ \\
Max gradient norm & -- & $0.5$ \\
Target KL divergence & $\delta_{\mathrm{KL}}$ & $0.01$ \\
Conjugate-gradient iterations & -- & $15$ \\
Conjugate-gradient damping & -- & $0.1$ \\
FVP sample frequency & -- & $4$ \\
Line-search steps & -- & $15$ \\
Line-search decay & -- & $0.8$ \\
Network architecture & -- & 2-layer, $[64,64]$ units \\
Network architecture (Seaquest) & -- & Nature CNN \citep{mnih2015human} \\
Initial log standard deviation & $\log\sigma_0$ & $0.0$ \\
Cost limit & $d$ & $0.01$ \\
Cost discount factor & $\gamma_c$ & $0.99$ \\
Cost GAE lambda & $\lambda_c$ & $0.95$ \\
\bottomrule
\end{tabular}
\end{table}

\begin{table}[H]
\centering
\caption{MASA PPO-Lag hyperparameters.}
\label{tab:PPOLag-hyperparameters}
\begin{tabular}{l c c}
\toprule
\textbf{Name} & \textbf{Symbol} & \textbf{Value}\\
\midrule
Learning rate & $\alpha$ & $1\times10^{-4}$ \\
Rollout steps & -- & $20{,}000$ \\
Batch size & $B$ & $256$ \\
PPO epochs & -- & $40$ \\
Discount factor & $\gamma$ & $0.99$ \\
GAE lambda & $\lambda$ & $0.95$ \\
Clip range & $\epsilon_{\text{clip}}$ & $0.2$ \\
Normalize reward advantages & -- & \texttt{True} \\
Normalize cost advantages & -- & \texttt{True} \\
Entropy coefficient & $\beta_{\mathrm{ent}}$ & $0.0$ \\
Value-loss coefficient & $\beta_V$ & $1.0$ \\
Max gradient norm & -- & $0.5$ \\
Network architecture & -- & 2-layer, $[64,64]$ units \\
Network architecture (Seaquest) & -- & Nature CNN \citep{mnih2015human} \\
Initial log standard deviation & $\log\sigma_0$ & $0.0$ \\
Cost limit & $d$ & $0.01$ \\
Cost discount factor & $\gamma_c$ & $0.99$ \\
Cost GAE lambda & $\lambda_c$ & $0.95$ \\
Initial Lagrange multiplier & $\lambda_{\mathrm{init}}$ & $10.0$ \\
Lagrange multiplier learning rate & $\alpha_\lambda$ & $0.1$ \\
Lagrange multiplier upper bound & $\lambda_{\max}$ & $100.0$ \\
\bottomrule
\end{tabular}
\end{table}

\section{Additional Diagnostics and Learning Curves}
\label{sec:additionaldiagnostics}
\begin{figure}[H]
    \centering
    \includegraphics[width=0.95\linewidth]{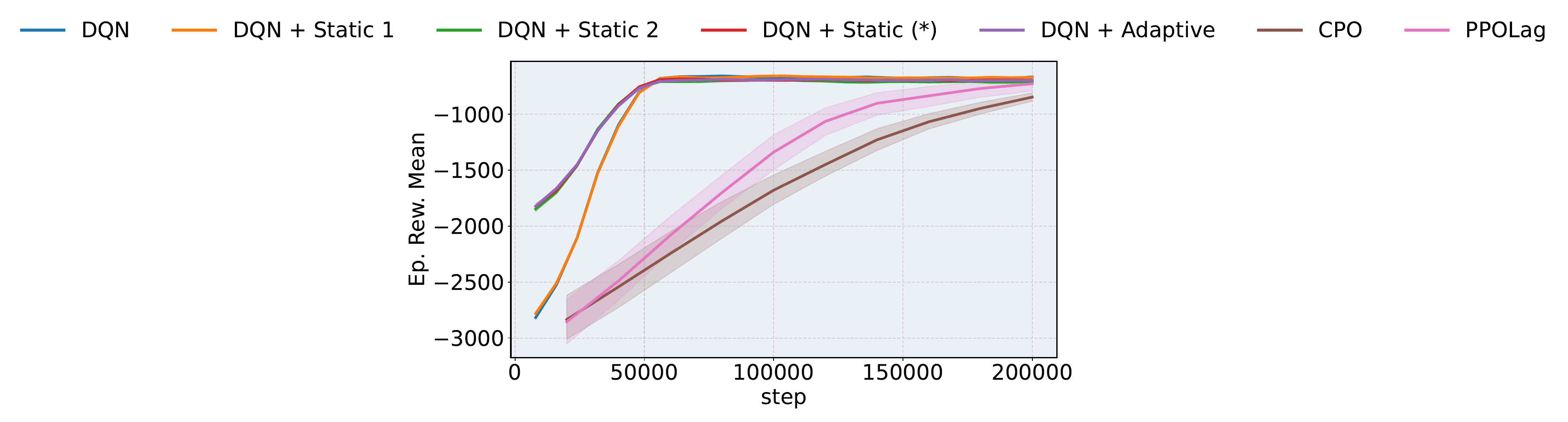}
    \caption{Learning curves for the Minepump case study.}
    \label{fig:learningcurves1}
\end{figure}

\begin{figure}[H]
    \centering
    \includegraphics[width=0.95\linewidth]{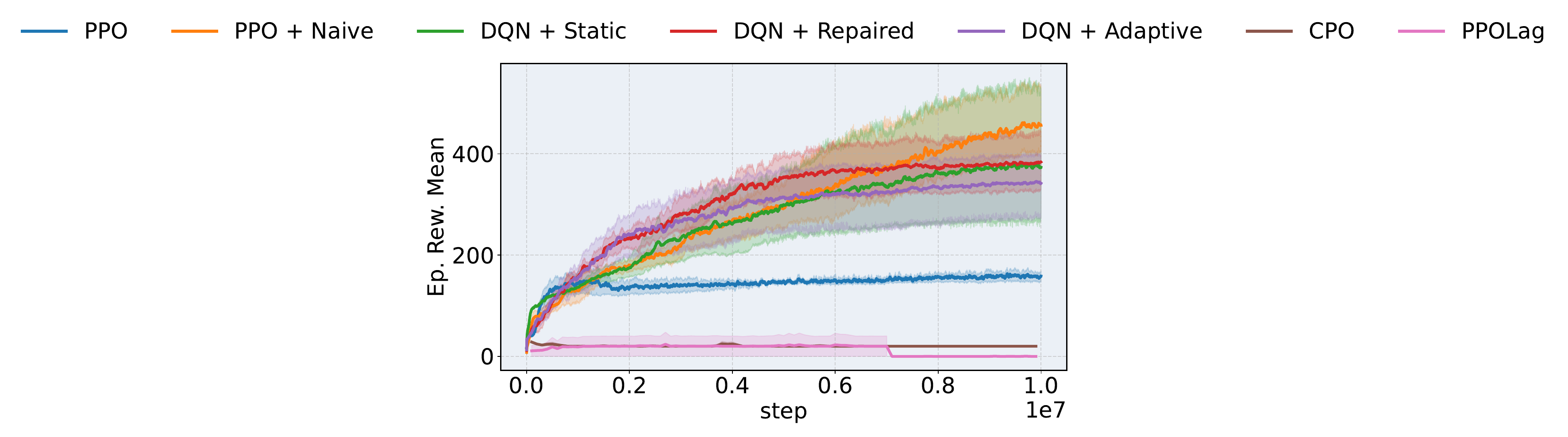}
    \caption{Learning curves for the Seaquest case study.}
    \label{fig:learningcurves2}
\end{figure}

In this section we provide the learning curves for both the Minepump and Seaquest experiments (c.f., Fig~\ref{fig:learningcurves1} and Fig.~\ref{fig:learningcurves2}).

\paragraph{Discussion.} Interestingly for the Minepump experiments there are two distinct types of learning dynamics among the algorithms. The first group: DQN and DQN + \textbf{Static 2}, converge much slower than the second group: DQN \textbf{Static 1}, \textbf{Static (*)} and DQN + \textbf{Adaptive}. The only difference here is that the second group all strictly enforce \textbf{guarantee1}, i.e., $G\big(\texttt{highwater}\to X(\texttt{pump})\big)$. This strict adherence to \textbf{guarantee1} acts as a sort of inductive bias that appears to speed up convergence, and interesting insight that demonstrates if the spec is well-aligned with the reward objective then learning can be sped-up. A similar picture is painted in the Seaquest learning curves: all algorithms that are constrained by, even a suboptimal shield during learning, achieve better performance and converge more quickly than the unconstrained RL baseline. 

\paragraph{Additional diagnostics.} We present some further diagnostics for the Minepump and Seaquest experiments. In particular for the Minepump example we log the average percentage (\%) of time spent in the winning region $\mathcal{W}$ during each execution trace, that is, states without deadlocks (where we chose an invalid action). Furthermore, we log the percentage of traces that satisfy \textbf{guarantee1} and \textbf{guarantee2} (c.f., Tab.~\ref{tab:minepump-diagnostics}). For the Seaquest experiments we also log the (\%) of time spent in the winning region $\mathcal{W}$ during each execution trace, and the percentage (\%) of time spent satisfying \textbf{guarantee1} (non-zero oxygen) during each execution trace (c.f., Tab.~\ref{tab:seaquest-diagnostics}).

\begin{table}[H]
\tiny
\centering
\caption{Additional diagnostics for the Minepump results over 10 seeds (mean $\pm$ SE).}
\label{tab:minepump-diagnostics}
\begin{tabular}{lcccccc}
\toprule
\textbf{Controller / Agent} & 
\textbf{Tr. \% in $\mathcal{W}$} & 
\textbf{Tr. \textbf{guar.~1}} & 
\textbf{Tr. \textbf{guar.~2}} & 
\textbf{Ev. \% in $\mathcal{W}$} & 
\textbf{Ev. \textbf{guar.~1}} & 
\textbf{Ev. \textbf{guar.~2}} \\
\midrule
Static Ctrl.~1  & 1.00  & 1.00 & 1.00 & 0.96 & 1.00 & 0.96 \\
Static Ctrl.~2  & 1.00  & 1.00 & 1.00 & 1.00 & 1.00 & 1.00 \\
DQN & 0.94 $\pm$ 0.00 & 0.03 $\pm$ 0.00 & 0.03 $\pm$ 0.00 & 0.95 $\pm$ 0.00 & 0.09 $\pm$ 0.05 & 0.00 $\pm$ 0.00 \\
DQN + \textbf{Static 1} & 0.94 $\pm$ 0.00 & 1.00 $\pm$ 0.00 & 0.00 $\pm$ 0.00 & 0.95 $\pm$ 0.00 & 1.00 $\pm$ 0.00 & 0.00 $\pm$ 0.00 \\
DQN + \textbf{Static 2}& 0.98 $\pm$ 0.00 & 0.00 $\pm$ 0.00 & 1.00 $\pm$ 0.00 & 0.98 $\pm$ 0.01 & 0.00 $\pm$ 0.00 & 1.00 $\pm$ 0.00 \\
DQN + \textbf{Static (*)}& 1.00 $\pm$ 0.00 & 1.00 $\pm$ 0.00 & 1.00 $\pm$ 0.00 & 0.99 $\pm$ 0.00 & 1.00 $\pm$ 0.00 & 0.00 $\pm$ 0.00 \\
DQN + \textbf{Adaptive (ours)} & 1.00 $\pm$ 0.00 & 1.00 $\pm$ 0.00 & 1.00 $\pm$ 0.00 & 1.00 $\pm$ 0.00 & 1.00 $\pm$ 0.00 & 1.00 $\pm$ 0.00 \\
CPO & 1.00 $\pm$ 0.00 & 1.00 $\pm$ 1.00 & 1.00 $\pm$ 0.00 & 0.93 $\pm$ 0.02 & 0.00 $\pm$ 0.00 & 0.00 $\pm$ 0.00  \\
PPOLag & 1.00 $\pm$ 0.00 & 1.00 $\pm$ 1.00 & 1.00 $\pm$ 0.00 & 0.94 $\pm$ 0.06 & 0.01 $\pm$ 0.00 & 0.00 $\pm$ 0.00  \\
\bottomrule
\end{tabular}
\end{table}

\begin{table}[H]
\tiny
\centering
\caption{Additional diagnostics for the Atari Seaquest results over 10 seeds (mean $\pm$ SE).}
\label{tab:seaquest-diagnostics}
\begin{tabular}{lcccc}
\toprule
\textbf{Controller / Agent} & 
\textbf{Train \% in $\mathcal{W}$} & 
\textbf{Train \% \textbf{guarantee1}} & 
\textbf{Eval \% in $\mathcal{W}$} & 
\textbf{Eval  \% \textbf{guarantee1}} \\
\midrule
PPO (Schulman et al. 2017)       &  0.74 $\pm$ 0.00 & 0.98 $\pm$ 0.00 & 0.83 $\pm$ 0.01 & 0.98 $\pm$ 0.00 \\
PPO + \textbf{Naive}       &  1.00 $\pm$ 0.00 & 1.00 $\pm$ 0.00 & 0.71 $\pm$ 0.02 & 0.99 $\pm$ 0.00 \\
PPO + \textbf{Static} (Alshiekh et al.)        &  1.00 $\pm$ 0.00 & 1.00 $\pm$ 0.00 & 0.71 $\pm$ 0.02 & 0.98 $\pm$ 0.00 \\
PPO + \textbf{Repaired} & 1.00 $\pm$ 0.00 & 1.00 $\pm$ 0.00 & 1.00 $\pm$ 0.00 & 1.00 $\pm$ 0.00 \\
PPO + \textbf{Adaptive (ours)}  & 1.00 $\pm$ 0.00 & 1.00 $\pm$ 0.00 & 1.00 $\pm$ 0.00 & 1.00 $\pm$ 0.00 \\
CPO & 1.00 $\pm$ 0.00 & 1.00 $\pm$ 0.00 & 1.00 $\pm$ 0.00 & 1.00 $\pm$ 0.00 \\
PPOLag & 1.00 $\pm$ 0.00 & 1.00 $\pm$ 0.00 & 1.00 $\pm$ 0.00 & 1.00 $\pm$ 0.00 \\
\bottomrule
\end{tabular}
\end{table}

\paragraph{Discussion (Minepump).} As expected DQN + \textbf{Static 1} satisfies \textbf{guarantee1} on 100\% of traces, similarly DQN + \textbf{Static 2} satisfies \textbf{guarantee2} on 100\% of traces. In the training environment DQN + \textbf{Static (*)} and DQN + \textbf{Adaptive} (including both static symbolic controllers) satisfy both guarantees on 100\% of traces and spend on 100\% of the time in the winning region. In the evaluation environment \textbf{Static Symbolic Controller 2} maintains correctness (as is expected), as does our DQN + \textbf{Adaptive}. Interestingly DQN + \textbf{Static (*)} satisfies \textbf{guarantee1} on all traces, indicating this is reward optimal behaviour, although of course fails to enforce \textbf{guarantee2}. DQN + \textbf{Static 1} and DQN + \textbf{Static 2} behave as expected here.

\paragraph{Discussion (Atari Seaquest).} In the training environment all shielded PPO agents maintain \textbf{guarantee1} and spend 100\% of the time in the winning region. The unconstrained PPO spends only 74\% of the time in the winning region, meaning at least 25\% of the time it is essentially doomed and will inevitably run out of oxygen.  As expected, \textbf{guarantee1} is only maintained by the pre-emptively repaired shield and the adaptive shield. Interestingly, the PPO + \textbf{Naive} and PPO + \textbf{Static} spend more time outside of the winning region than the unconstrained PPO agent here, probably because these two agents are more competent and don't get prematurely hit by an enemy.

\subsection{Access to Code}

Our code is made publicly accessible via GitHub: \url{https://github.com/sacktock/Adaptive_GR-1-_Shielding}

\section{Atari Seaquest Spec (full)}
\label{sec:seaquestspec}

First we describe all Boolean variables $\mathcal{V} = \mathcal{X}_1 \cup \mathcal{X}_2 \cup \mathcal{Y}$.
\begin{itemize}
\item \textbf{Environment variables $\mathcal{X}_1$:} \texttt{oxygen0}, \texttt{oxygen1}, \texttt{oxygen2}, \texttt{oxygen3}, \texttt{oxygen4}, \texttt{oxygen5}, \texttt{oxygen6} (7 oxygen bits for encoding 64 different oxygen values), \texttt{depth0}, \texttt{depth1}, \texttt{depth2}, \texttt{depth3}, \texttt{depth4} (5 depth bits encoding 24 different depth levels, corresponding to the multiples of 4 from 0-92 inclusive), \texttt{diver\_at\_depth1}, \texttt{diver\_at\_depth2}, \texttt{diver\_at\_depth3}, \texttt{diver\_at\_depth4} (environment variables that are asserted when there is a diver at each of the 4 possible levels).

\item \textbf{Auxiliary environment variables $\mathcal{X}_2$:} \texttt{operational} (whether the submarine is operational; able to move up or down), \texttt{reset\_flag} (specifies if the system has just been reset or not); \texttt{faster\_rate} (whether the oxygen is depleting at a faster rate or not)
    
\item \textbf{System variables $\mathcal{Y}$:} \texttt{up} (agent actions corresponding to upward movements), \texttt{down} (agent actions corresponding to downward movements), \texttt{rescues1}, \texttt{rescues2}, \texttt{rescues3}, \texttt{rescues4} (system variables that are asserted to rescue a diver at each of 4 the possible levels).
\end{itemize}
Now we describe the initial conditions $\theta^{\mathbf e}$ and $\theta^{\mathbf s}$ of  the environment $\mathbf e$ and system $\mathbf s$ respectively.
\begin{itemize}
    \item \textbf{Environment $\theta^{\mathbf e}$:} $\texttt{depth}=0$ (encoded via \texttt{depth0}-\texttt{depth4}), $\texttt{oxygen}=0$ (encoded via \texttt{oxygen0}-\texttt{oxygen6}),  $\texttt{diver\_at\_depth1}$-$ \texttt{diver\_at\_depth4} = False$ (no divers present initially), $\texttt{operational}=False$ (the agent cannot move at the start), $\texttt{reset\_flag}=True$, $\texttt{faster\_rate} = False$.
    \item \textbf{System $\theta^{\mathbf s}$:} $\texttt{up} = False$ and $\texttt{down} = False$ (the submarine is initially stationary), $\texttt{rescues1}$-$\texttt{rescues4} = False$ (no divers initially rescued).
\end{itemize}
Now we describe the environment invariants $\rho^{\mathbf e}$.
\begin{itemize}
    
    \item \textbf{assumption1} -- oxygen does not increase until \texttt{reset\_flag} is true: $G((\texttt{oxygen}=0 \land \texttt{reset\_flag}) \to X(\texttt{oxygen=0}))$
    \item \textbf{assumption2} -- oxygen increases on the surface after reset\_flag is false:
    $G((\texttt{oxygen} = 0\land \texttt{depth} =0 \land \neg \texttt{reset\_flag}) \to X(\texttt{oxygen}=1))$
    \item \textbf{assumption3} -- \texttt{operational} when oxygen is at a maximum: $G((\texttt{oxygen} = 64) \to \texttt{operational})$.
    \item \textbf{assumption4} -- \texttt{operational} when below the surface: $G ((\texttt{depth}> 0) \to \texttt{operational})$
    \item \textbf{assumption5} -- not \texttt{operational} after we resurface until we have fully replenished our oxygen supplies: $G((\texttt{depth}=4) \land \texttt{up} \to X(\texttt{depth} = 0 \land \neg \texttt{operational}))$
    \item \textbf{assumption6} -- oxygen depletion rate: $G((\neg \texttt{faster\_rate} \land \texttt{depth} > 0 \land \texttt{oxygen} > 0) \to X(\texttt{oxygen}) = \texttt{oxygen} -1)$. This can be encoded via binary variables, for example,
    \begin{align*}
        &G(\neg \texttt{faster\_rate} \land (\texttt{depth0} \lor \texttt{depth1} \lor \ldots \lor \texttt{depth4}) \land (\texttt{oxygen0} \land \texttt{oxygen1} \land \texttt{oxygen2} \land \\ & \quad \ldots \land \texttt{oxygen6} \land \neg \texttt{oxygen7})
         \to X( (\neg \texttt{oxygen0} \land \texttt{oxygen1} \land \texttt{oxygen2} \land \ldots \land \texttt{oxygen6} \land \neg \texttt{oxygen7})))
    \end{align*}
    i.e., the oxygen level goes from 63 to 62. Similarly when $\texttt{faster\_rate}=True$,
    \begin{align*}
        G((\texttt{faster\_rate} \land \texttt{depth} > 0 \land \texttt{oxygen} > 0) \to X(\texttt{oxygen}) = \texttt{oxygen} -2)
    \end{align*}
    i.e., the oxygen depletes at a faster rate of $2$ units per timestep.
    \item \textbf{assumption7} -- \texttt{faster\_rate} does not turn on at the surface: $G((\texttt{depth} = 0) \to X(\neg\texttt{faster\_rate}))$
    \item \textbf{assumption8} -- once \texttt{faster\_rate} is triggered it never turns off: $G(\texttt{faster\_rate} \to X(\texttt{faster\_rate}))$
    \item \textbf{assumption9} -- \texttt{faster\_rate} does not turn on at any oxygen level: $G((\texttt{oxygen} = i) \to X(\neg\texttt{faster\_rate}))$. This is encoded via binarization for every oxygen level.
    \item \textbf{assumption10} -- movement of the submarine: $G((\texttt{depth}>0) \land \texttt{up} \to X(\texttt{depth})=\texttt{depth} - 4)$ (i.e., an $\texttt{up}$ movement) can also be encoded via binary variables, for example,
    \begin{equation*}
        G((\texttt{depth0} \land \texttt{depth1} \land \ldots \land \texttt{depth3}\land \neg \texttt{depth4}) \land \texttt{up} \to X (\neg\texttt{depth0} \land \texttt{depth1} \land \ldots \land \texttt{depth3}\land \neg \texttt{depth4}) )
    \end{equation*}
    i.e., the depth decreases from 15 to 14 when $\texttt{up}=True$.
    \item \textbf{assumption11} -- movement of the submarine but for $\texttt{down}$: $G((\texttt{depth}<92) \land \texttt{down} \to X(\texttt{depth})=\texttt{depth} + 4)$. For brevity we omit any further binarization here. 
    \item \textbf{assumption12} -- stationarity of the submarine when neither $\texttt{up}=True$ or $\texttt{down}=True$: $G((\neg \texttt{up} \lor \neg \texttt{down}) \to X( \texttt{depth})=\texttt{depth})$
    \item \textbf{assumption13} -- oxygen increases at the surface: $G((\texttt{depth} = 0 \land \texttt{oxygen} < 64) \to X(\texttt{oxygen}) = \texttt{oxygen} + 1)$. 
    \item \textbf{assumption14} -- diver is rescued if the submarine is present at the corresponding depth. Noting there are 4 possible locations for the diver to be present which divide the depth levels evenly. Thus, we have the following: 
    \begin{itemize}
        \item $G((\texttt{depth}=24 \land \texttt{diver\_at\_depth1}) \to X (\neg \texttt{diver\_at\_depth1} ))$
        \item $G((\texttt{depth}=44 \land \texttt{diver\_at\_depth2}) \to X (\neg \texttt{diver\_at\_depth2} ))$
        \item $G((\texttt{depth}=68 \land \texttt{diver\_at\_depth3}) \to X (\neg \texttt{diver\_at\_depth3} ))$
        \item $G((\texttt{depth}=92 \land \texttt{diver\_at\_depth4}) \to X (\neg \texttt{diver\_at\_depth4} ))$
    \end{itemize}
    i.e., for each diver depth level.
    \item \textbf{assumption15} -- persistent diver: $G((\texttt{diver\_at\_depth1} \land \neg \texttt{depth}=24)\to X(\texttt{diver\_at\_depth1}))$ (diver does not disappear unless rescued), defined analogously for \texttt{diver\_at\_depth1}-\texttt{diver\_at\_depth4}.
\end{itemize}
And the environment justice goals $J_i^{\mathbf e}$.
\begin{itemize}
    \item \textbf{assumption16} -- \texttt{reset\_flag} must always eventually be turned off: $GF(\neg \texttt{reset\_flag})$.
    \item \textbf{assumption17} -- the environment must always eventually present the system with divers. Formally this is written as:
    \begin{itemize}
        \item $GF(\texttt{diver\_at\_depth1})$.
        \item $GF(\texttt{diver\_at\_depth2})$.
        \item $GF(\texttt{diver\_at\_depth3})$.
        \item $GF(\texttt{diver\_at\_depth4})$.
    \end{itemize}
\end{itemize}
Now we describe the system invariants $\rho^{\mathbf s}$.
\begin{itemize}
    \item \textbf{guarantee0} -- we can never move up and down simultaneously: $G(\neg (\texttt{up} \land \texttt{down}))$.
    \item \textbf{guarantee1} -- never run out of oxygen: we don't have to explicitly write ``never run out of oxygen'', i.e., $G(\neg(\texttt{oxygen} = 0))$. This was an oversimplification written in the main text. Rather, we just restrict the movement of the submarine when $\texttt{oxygen} = 0$. If the submarine is not operational then it can't pursue all its goals, so any state with $\texttt{oxygen} = 0$ would lie outside the winning region $\mathcal{W}$ anyway, thus we write (in binary form),
    \begin{equation*}
        G((\neg \texttt{oxygen}1 \land \ldots \land\neg \texttt{oxygen7}) \lor \texttt{operational} \to (\neg \texttt{up} \land \neg\texttt{down}))
    \end{equation*}
\end{itemize}
Finally the system goals $J_i^{\mathbf s}$,
\begin{itemize}
    \item  \textbf{guarantee2} -- rescue divers at a given depth segment: \\ $G(\texttt{diver\_at\_depth}[i] \to F(\texttt{depth} =i))$, i.e., ensures we always eventually reach divers at a given depth. Formally we write,
    \begin{itemize}
        \item $GF(\texttt{rescues1})$ (for $\texttt{rescues1}$-\texttt{rescues4}).
        \item $G((Y(\texttt{diver\_at\_depth1}) \land \texttt{diver\_at\_depth1}) \to \texttt{rescues1})$ (for $\texttt{diver\_at\_depth1}$-\texttt{diver\_at\_depth4}  and $\texttt{rescues1}$-\texttt{rescues4} analogously). This is technically a system invariant $\rho^{\mathbf s}$ enforcing the $\texttt{rescues1}$ flag can only be true when a diver has disappeared.
        \item Further, $G(\texttt{diver\_at\_depth1} \to \neg \texttt{rescues1})$ (for $\texttt{diver\_at\_depth1}$-\texttt{diver\_at\_depth4}  and $\texttt{rescues1}$-\texttt{rescues4} analogously) asserts the $\texttt{rescues1}$ flag cannot be on when there is a diver to be rescued (also a system invariant $\rho^{\mathbf s}$).
    \end{itemize}
\end{itemize}

\end{document}